# 3D Scene Geometry-Aware Constraint for Camera Localization with Deep Learning

Mi Tian[+], Qiong Nie[+], Hao Shen[*]

*Abstract*— Camera localization is a fundamental and key component of autonomous driving vehicles and mobile robots to localize themselves globally for further environment perception, path planning and motion control. Recently end-to-end approaches based on convolutional neural network have been much studied to achieve or even exceed 3D-geometry based traditional methods. In this work, we propose a compact network for absolute camera pose regression. Inspired from those traditional methods, a 3D scene geometry-aware constraint is also introduced by exploiting all available information including motion, depth and image contents. We add this constraint as a regularization term to our proposed network by defining a pixel-level photometric loss and an image-level structural similarity loss. To benchmark our method, different challenging scenes including indoor and outdoor environment are tested with our proposed approach and state-of-the-arts. And the experimental results demonstrate significant performance improvement of our method on both prediction accuracy and convergence efficiency.

## I. Introduction

Camera localization, as a foundation for many applications such as autonomous driving vehicle and mobile robots, estimates camera position and orientation from a query image and a pre-built map with scene information. In traditional localization framework, this scene information is generally presented as sparse key points with 3D information and feature descriptor. Camera poses are then estimated from 2D-3D matching between query images and a map by applying a Perspective-n-Point (PnP) solver accompanied with RANSAC [16, 38] strategies for outlier removal. Different methods are proposed to improve efficiency and effectiveness of such 2D-3D matching. For instance, image-level features like bag-of-words [32, 33], VLAD [34], Fish Vector [36, 37] are usually employed for similarity matching between query images and keyframes stored during mapping. Due to the image-level features retrieval results, matching area can be reduced into top N most similar keyframes and their surrounding points, which means that only a small 3D submap will participate in 2D-3D matching. As an intermediate step, these utilizing keyframes retrieval are categorized into retrieval-based approaches [3, 30, 31]. However direct approaches take advantages of different hashing algorithms to match 2D-3D points for computation acceleration. Specifically, bag-of-words [32, 33] and LSH [23] are two popular hashing methods for camera localization. Although many different efforts are made to improve 2D-3D matching accuracy, the fact that traditional approaches are based on low-level features such as SIFT [11, 1], SURF [25, 9], ORB [18], etc. makes it difficult to deal with challenging

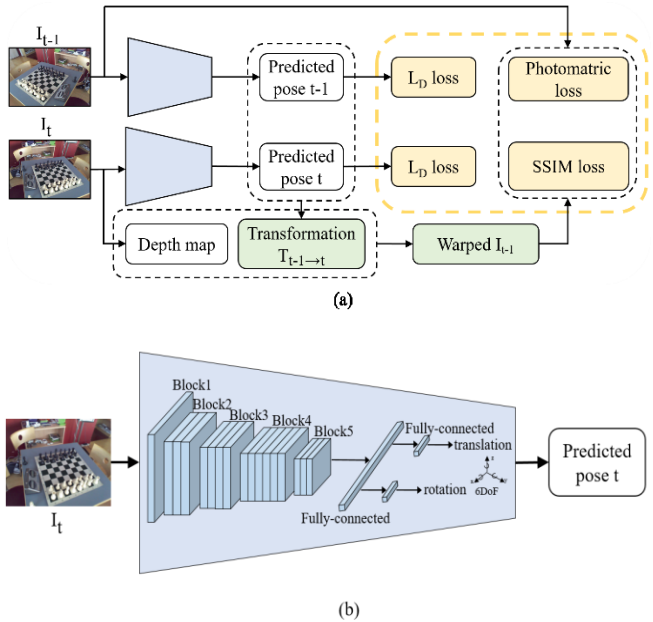

Figure 1: Schematic representation of our proposed self-supervised deep learning for camera localization with 3D scene geometry-aware constraint. (a): Training flow of our proposed algorithm which requires a pair of RGB images and a depth map of one of them. Green rectangles are computational components according to predicted poses and depth map without learnable parameters. Blue rectangles are networks for pose regression to be trained. And yellow rectangles are constraint terms of network. (b): Inference flow of camera pose localization. Blue part is network architecture based on the ResNet-50 that is a detailed description for the blue part in (a).

environments like illumination change or seasonal change.

While learning-based methods aim to regress 6 DoF pose in an end-to-end way [5, 6, 17]. Scene information in this case is described as neural network weights and mapping step turns into a network training process. The first deep learning framework PoseNet [2] retrieves camera pose from a single image. [29] exploits temporal information for pose estimation by utilizing image sequence. [15] introduces the encoder-decoder architecture into camera localization. Some other changes like reasoning about the uncertainty of the estimated poses [22] are also proposed. However, all these methods train their networks by a naive Euclidean distance between prediction and ground truth pose. Inspired from traditional methods utilizing 3D geometry information, recently many geometry relevant loss functions such as geometric consistency error [7, 8], reprojection error [4], relative transform error [24] are built as regularization terms. Such

[+] indicates equal contributions. [*] indicates corresponding author. Email: ({tianmi02, nieqiong, shenhao04}@meituan.com).

This research is supported by Beijing Science and Technology Project (No. Z181100008918018). All authors are with Meituan-Dianping Group, Beijing, China.

methods perform better than those learned from single image information.

We follow prior works of learning-based camera localization and further search for more geometric information to constraint our model. In addition to standard sensors like GPS and camera that usually provide ground truth poses and images for localization, depth sensors are also very popular in SLAM applications. For indoor situation, we can directly obtain depth information from structured light camera, time-of-flight camera or stereo camera with available depth estimation algorithm. For outdoor environment, 3D LIDAR is usually employed for both localization and scene perception. From 3D geometry knowledge, when a general point in 3D scene is viewed in several images, their corresponding pixel intensities are supposed to be identical. This property we called as photometric consistency. It is the base idea for many direct visual odometry methods [3, 19] or SLAM methods [20, 32 - 35].

In this paper, we immigrate this idea into a neural network. The photometric consistency is described as a photometric loss term accompanied with a structured similarity SSIM [10] loss function to optimize pose regression with self-supervised learning. Meanwhile ground truth pose information and depth information (sparse or dense) from whatever depth sensors are used during training process only to calculate the photometric error loss. It bootstraps the loss function by penalizing pose predictions that contradict 3D scene geometry and helps the convergence of network. Although many traditional stereo methods and learning-based methods can estimate depth information, we prefer to use ground truth depth captured by robust sensors, considering easy availability of the sensor and information accuracy, and also our method does work even with very sparse depth information.

To this end, we make the following contributions compared to other works: (i) We propose a deep neural network architecture to directly estimate an absolute camera pose from an input image. (ii) By utilizing depth sensor information, we applied an additional 3D scene geometry-aware constraint to improve prediction accuracy. As mentioned, sparse depth information will be enough to get remarkable localization precision increment. This means that our method can be adapted with any kind of depth sensors (sparse or dense). (iii) We present extensive experimental evaluations on both indoor and outdoor datasets to compare our approach with state-of-the-art methods. At the same time, we demonstrate that the proposed additional 3D scene geometry-aware constraint can be easily added into other network and make performance improvement.

## II. RELATED WORK

Various CNN-based approaches of absolute camera localization have been proposed in the literature. In this section, some of the techniques developed thus far for improving the performance of localization will be discussed.

**CNN-based camera localization** was first proposed by PoseNet [2] which utilized base architecture of GoogLeNet to directly regress 6DoF camera pose with an input RGB image. By using Bayesian CNN, the authors extended their work to model precision uncertainty [22]. Following approaches, mainly differ in underlying base architecture and loss function used for training. Melekhov et al. [15] proposed Hourglass Network described as a symmetric encoder-decoder structure, which is widely used for applications of semantic segmentation. Rather than using a single image, Walch et al. [13] and Xue et al. [14] introduced Long-Short Term Memory (LSTM) to exploit global information by features learning from constraint of temporal smoothness of the video stream. Valada et al. [7, 8] proposed multitask learning framework for visual localization, odometry estimation and semantic segmentation. This method, which exploits inter-dependencies within multitask for the mutual benefit of each task, is considered as state-of-the-art since it provides higher localization precision than many other CNN-based approaches. However, such multitask training process requires much ground truth information, especially labeled semantic segmentation data causing this approach not flexible in many application domains.

**Geometric consistency Constraint** is recently used to help improving accuracy of pose regression and proved more effective than that of using Euclidean distance constraint alone. Valada et al. [7, 8] introduced geometric consistency to bootstrap loss function by penalizing pose predictions that contradict the relative motion. MapNet [24] imposed a constraint on relative pose between image pairs for global consistency. This method provided stricter constraints without any additional input information required as relative pose is easily computable by absolute ground truth pose. Kendall et al. [4] introduced another geometric loss named reprojection error defined as the residual of 3D points projected onto 2D image plane using the ground truth and predicted pose. All these works are considered to be state-of-the-art of that time using geometry consistency loss. In our work, we explore a 3D scene geometry-aware constraint called photometric error constraint. 3D structure information is added into this constraint which enforces network not only align predicted poses to camera motion but also aggregate scene structure model. Compared with the above image-level geometry consistency losses, our method makes use of geometry information of every 3D point of the scene and provides much stronger pixel-level constraint.

**Photometric error constraint** is typically used to deal with relative pose regression, optical flow estimation and depth prediction with supervised or unsupervised learning. For instance, Ma et al. [27] explored temporal relations of video sequences to provide additional photometric supervisions for depth completion network. Zhou et al. [12] built CNNs with unsupervised learning of dense depth and camera pose with photometric error loss to learn the scene level consistent movement governed by camera motion. Yin et al. [26] proposed a multitask unsupervised learning method of dense depth, optical flow and egomotion prediction, where photometric error constraint played an important role to enforce consistency between different tasks. Shen et al. [28] proposed to bridge the gap between geometric loss and photometric loss by introducing the matching loss constrained by epipolar geometry. Since photometric error constraint has been proved effective for relative pose regression and depth prediction, we introduce this photometric error constraint and validate its effectiveness on absolute pose prediction. As our knowledge, this is the first time that photometric error is imposed to solve absolute pose regression problem.

## III. PROPOSED APPROACH

Our method is dedicated to absolute pose regression. The ground truth pose and depth information will be used during training process. Both information are easily available from sensors like GPS and depth sensors like RGBD cameras or LIDARs. At any inference time, only one image is imported to the network to localize the camera itself. In this section, we will introduce our pose regressing neural network as first. Then we will explain both training and inference framework in detail. At training process, three constraints are applied to help learning process towards a global minimum: a classic Euclidean error to measure distance from prediction to ground truth pose as well as two regularization terms formulated as a photometric loss and a structural similarity loss. Both regularizations try to lead model to obey photometric consistency but respectively by pixel-level and image-level. Finally, a warping process which is an intermediate step for building both terms is also presented.

### A. Network architecture

We build a CNN architecture to predict the corresponding absolute pose p = [x, q] for a given image, where x denotes position and q denotes a unit of quaternion representing orientation. We use the first five residual blocks of ResNet-50 as backbone and modify it by introducing a global average pooling layer after the last residual block, and subsequently add three fully connected layers with 2048 neurons, 3 neurons and 4 neurons respectively. The last two fully connected layers separately output the absolute position x and orientation q (see Figure 1(b)). Each convolution layer is followed by batch normalization and Rectified Linear Unit (ReLU)

At inference process, only current image is applied to the network for regressing 6DoF pose directly (see Figure 1(b)). While during training (see Figure 1(a)), two successive images $I_{t-1}$ and $I_t$ as well as a depth map of $I_{t-1}$ and the corresponding ground truth poses of $I_{t-1}$ and $I_t$ are required. The network learns weights and predicts absolute pose for both images by building Euclidean distant constraint as a loss term for each prediction. For a moving camera, two consecutive images are usually overlapped and their absolute poses can be mutually constrained by 3D scene geometry. In this paper, this 3D scene geometry-aware constraint is described as photometric error and SSIM error. Compared to [24] which just employs relative transform as geometry constraint to learn absolute pose, in our work, 3D scene geometry-aware constraint is employed as a pixel-level loss, exploiting more information including relative transform, 3D information and pixel intensity to learn camera localization with a global optimization directly and efficiently.

### B. Warping computation

The warping computation from image $I_{t-1}$ to $I_t$ is illustrated in the following:

$$u_t = K T_{t-1}^t D_{t-1}(p_{t-1}) K^{-1} u_{t-1} \quad (1)$$

Where $u_{t-1}$ is a static pixel in previous image $I_{t-1}$, its warped pixel to current time t is defined as $u_t$. We can easily get intrinsic matrix K by camera calibration. The 3D transform matrix from previous image to the current $T_{t-1}^t$ can be computed according to their absolute poses $T_w^{t-1}$ and $T_w^t$:

$$T_{t-1}^t = T_w^t * (T_w^{t-1})^{-1} \quad (2)$$

In warping computation, depth information $D_{t-1}(p_{t-1})$ is required for reconstructing 3D structure from 2D image pixels. As we explained in the previous section, dense depth information is not necessary. So we can extract it from depth sensors (structural light cameras, Time-of-flight cameras, stereo sensors and 3D LIDAR) or from stereo like depth computation algorithms, for example triangulation method of matched points from two overlapped images with knowing transform between them. However, to make sure not introducing extra depth error into our model. We prefer to choose robust depth information from a sensor.

To facilitate gradient computation for backpropagation, we create a synthetic image $warped_{t-1}$ with the same format of current image $I_t$ by using bilinear interpolation as sampling mechanism for warping. As the warping is fully differentiable, we do not need any pre-computation for training and online running. Furthermore, no learnable weight or additional overhead is required for training and inference.

### C. Loss function

In this section, constraint terms used for training network will be discussed in detail. In addition to typical Euclidean distant constraint, we introduce photometric loss term and structure similar loss term based on the warping results.

**Euclidean distant constraint** Since we input two successive images into the model in parallel during training, the Euclidean distant losses for both images are calculated as:

$$L_D = L_D(I_{t-1}) + L_D(I_t) \quad (3)$$

with

$$L_D(I_i) = \|x_i - \hat{x}_i\|_2 + \beta \|q_i - \hat{q}_i\|_2 \quad for\ i \in \{t-1, t\} \quad (4)$$

Where $x_i$, $q_i$ are the ground truth position and orientation, $\hat{x}_i, \hat{q}_i$ are the predicted position and orientation, and $\beta$ is a weighted parameter to keep the expected values of position and orientation errors to be nearly equal and to be trained online. This highly strong supervision signal leads pose prediction converge to the approximate ground truth.

**Photometric error constraint** When there is limited change of viewpoint and the environment is assumed to be light-invariant, the intensity values of a 3D point in different images are supposed to be the same. This photometric consistency is used for solving many problems (both traditional solution and learning-based solutions) like optical flow estimation, depth estimation, visual odometry, etc. Here, we employ it for absolute pose estimation. Here the loss function is designed as the difference between the $warped_{t-1}$ image and current image $I_t$:

$$L_P = \sum_{i,j} M(u_{t-1}^{i,j}) \|I_t(i,j) - warped_{t-1}(i,j)\|_1 \quad (5)$$

Where $u_{t-1}^{i,j}$ is the pixel with coordinate $(i,j)$ in image $I_{t-1}$, $M(u_{t-1}^{i,j})$ is an image mask. The idea is to mask pixels without depth information and that do not obey photometric consistency. In our case, we mainly use it to mask two types of pixels: moving pixels and pixels with invalid depth information. The depth validity depends on the acquisition methods. For instance, depth from range sensors like LIDAR usually has satisfactory accuracy even at a long distance, but depth from computation algorithms like stereo-like method is much noisy. And many strategies can be applied to remove dynamic objects. For example, as long as moving objects are

Table1: Comparison of median localization error with existing CNN-based models on 7-Scene dataset

| Scene | Spatial extent | PoseNet [2] | LSTM-Pose [13] | VidLoc [29] | Hourglass Pose[15] | PoseNet2 [4] | MapNet [24] | Ours |
|---|---|---|---|---|---|---|---|---|
| Chess | 3×2×1 m$^3$ | 0.32, 8.12° | 0.24, 5.77° | 0.18, N/A | 0.15, 6.53° | 0.13, 4.48° | 0.08, 3.25° | 0.09, 4.39° |
| Fire | 2.5×1×1 m$^3$ | 0.47, 14.4° | 0.34, 11.9° | 0.26, N/A | 0.27, 10.84° | 0.27, 11.3° | 0.27, 11.69° | 0.25, 10.79° |
| Heads | 2×0.5×1 m$^3$ | 0.29, 12.0° | 0.21, 13.7° | 0.14, N/A | 0.19, 11.63° | 0.17, 13.0° | 0.18, 13.25° | 0.14, 12.56° |
| Office | 2.5×2×1.5 m$^3$ | 0.48, 7.68° | 0.30, 8.08° | 0.26, N/A | 0.21, 8.48° | 0.19, 5.55° | 0.17, 5.15° | 0.17, 6.46° |
| Pumpkin | 2.5×2×1 m$^3$ | 0.47, 8.42° | 0.33, 7.00° | 0.36, N/A | 0.25, 7.01° | 0.26, 4.75° | 0.22, 4.02° | 0.19, 5.91° |
| RedKitchen | 4×3×1.5 m$^3$ | 0.59, 8.64° | 0.37, 8.83° | 0.31, N/A | 0.27, 10.15° | 0.23, 5.35° | 0.23, 4.93° | 0.21, 6.71° |
| Stairs | 2.5×2×1.5 m$^3$ | 0.47, 13.8° | 0.40, 13.7° | 0.26, N/A | 0.29, 12.46° | 0.35, 12.4° | 0.30, 12.08° | 0.26, 11.51° |

usually vehicles or persons, object detection can be applied in advance to remove these moving objects. Moreover, we can also ignore the pixels with large photometric errors since these pixels are suspectable to violate the consistency principle.

Minimizing the photometric error takes effect only when the warped pixel is very close to the true correspondence. It requires predicted pose not far from ground truth. At the early epochs of training, Euclidean loss determines the gradient direction dominantly as current predicted pose is very different from ground truth and therefore photometric loss produces only a weak or even bad effects. To this end, we propose a self-adaption strategy: a photometric error is used for back-propagation only when the projection point $u_{t-1}$ and $u_t$ satisfies $\|u_t - u_{t-1}\|_1 \leq h$ ($h$ is a threshold value that highly depends on scenes, in our case, $h$ is set as 10). The purpose is to maximize the value of photometric loss for optimizing pose prediction.

**Structural similarity constraint** This constraint tries to extract structural information from scene, like the way of human visual system. The similarity of two images I_x and I_y is formulated as:

$$SSIM(x,y) = \frac{(2\mu_x\mu_y + C_1)(2\sigma_{xy} + C_2)}{(\mu_x^2 + \mu_y^2 + C_1)(\sigma_x^2 + \sigma_y^2 + C_2)} \quad (6)$$

Where $C_1$ and $C_2$ are constant to keep SSIM valid. The SSIM value is [0,1] and high similarity corresponds to a big value. An auxiliary constraint that differs $warped_{t-1}$ image and current image $I_t$ is defined by equation (7) in combination with photometric error.

$$L_S = \frac{1 - SSIM(I_t, warped_{t-1})}{2} \quad (7)$$

The final loss function is defined in formulate (8). It contains three loss terms with different weighted parameters namely $\lambda_D, \lambda_P, \lambda_S$ to balance every loss term, and constrain the weights update together. Both Euclidean distant loss term and SSIM loss term are image-level constraints, while photometric loss term belongs to a pixel-level constraint, which can lead to a more precise accuracy of prediction.

$$L = \lambda_D L_D + \lambda_P L_P + \lambda_S L_S \quad (8)$$

## IV. EXPERIMENT EVALUATION

In this section, we will present experimental results of our proposed method for camera localization in comparison with several state-of-the-art works both on indoor and outdoor datasets. The results demonstrate that our introduced loss terms as well as self-supervised strategy for absolute camera localization task are outstanding in prediction accuracy as well as training convergence.

*A. Datasets*

Our method is evaluated on a well-known public dataset – Microsoft 7-Scene which is a collection of tracked RGB-D camera frames [21]. Seven different scenes recorded from a handheld Kinect RGB-D camera at 640×480 resolution are proposed for evaluation. The dense depth map is directly obtained from RGB-D sensors and ground truth camera pose is provided by KinectFusion algorithm. The existence of motion blur and weak texture under office environment makes this 7-scene dataset very challenging and widely evaluated by localization and tracking algorithms. To facilitate comparison, we take the same training and testing sequence split of each scene as other methods did.

Oxford robotcar dataset [39] contains 100 repetitions of a consistent route through central oxford captured twice a week over a period of over a year. Different types of data are available from multiple sensors including monocular cameras, LIDAR, GPS, INS measurements as well as stereo cameras. We take sub-dataset LOOP with a total length of 1120m for our evaluation. Two subsets overlapping the whole path with the same motion direction are used for training and test respectively.

*B. Implementation details*

Since [7, 8] demonstrate that neither synthetic pose augmentation nor synthetic view augmentation techniques yield any performance gain. In some cases, they have even negative impacts on pose accuracy. In our experiments, we only take proven well-performed preprocessing steps like resize of input images into 320×240 and normalization.

We use the Adam solver for optimization with $\beta_1 = 0.9$, $\beta_2 = 0.99$, and $\varepsilon = 10^{-10}$. We initialize five residual blocks with weights of ResNet-50 pre-trained on ImageNet and remaining layers with Gaussian distribution, then fine-tuning all layers with mini-batch size of 12 and maximum iterations of 50 epochs. We apply layers-wise learning rate set that is initialized as 8e-4 and 2e-4 for five residual blocks and remaining layers respectively. Polynomial decay for learning rate is adopt with power = 0.9. The weighted parameters $\beta$, $\lambda_D$, $\lambda_P$, $\lambda_S$ are set as 3, 1, 0.01, 0.1 on all scenes. The work is implemented based on Tensorflow deep learning library and all the experiments are performed on a NVIDIA Titan V GPU with 16GB on-board memory.

*C. Comparison with prior methods*

Our regression method is tested on all scenes of 7-Scene dataset to compare with prior CNN-based methods namely

PoseNet [2], LSTM-Pose [13], VidLoc [29], Hourglass-Pose [15], PoseNet2 [4] and MapNet [24]. Table 1 shows the quantitative comparisons of median translation and rotation errors for each scene in the datasets. Except that MapNet slightly outperforms on chess scene, our method obtains better results on most scenes. Moreover, compared to MapNet [24] that needs 300 epochs and PoseNet [2] needs more, our method takes only 50 epochs iterations to convergence.

To illustrate our results in detail, several camera pose trajectories on test sequences of heads, fire, pumpkin and stairs scenes are shown in Figure 2. It is obvious that trajectories provided by PoseNet [2] are much noised and even fail sharply in some places. MapNet [24] has a stable prediction globally but the accuracy is unsatisfactory. In this experiment, our method achieves mostly outstanding performances both on translation and rotation accuracy. From above experiments, our proposed network architecture collaborated with introduced photometric error loss term exhibits much better performances considering accuracy-efficiency balance.

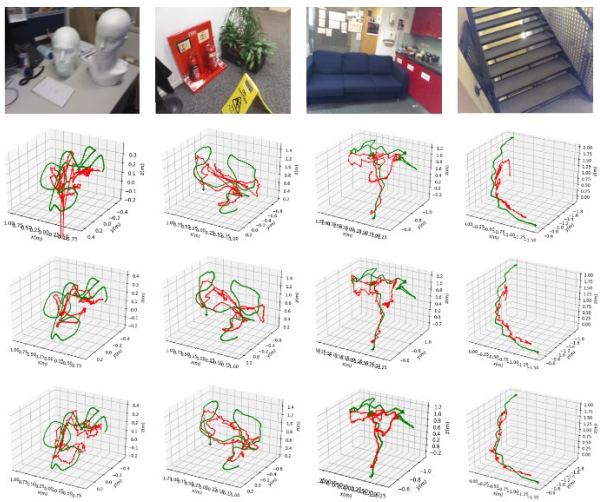

Figure 2: camera localization results on Microsoft 7-Scene. From left to right, the four test sequences are heads-01 sequence, fire-04 sequence, pumpkin-01 sequence and stairs-01 sequence. From top to bottom, the three results are from PoseNet [2], MapNet [24] and our method respectively (green for the ground truth, red for the prediction).

### D. Ablation studies

In this section, the performance of our proposed geometric constraint for the absolute camera localization task will be studied. To this end, we employ ablation experiments that we train different network architectures including GoogLeNet of PoseNet and ours with the help of photometric error loss and SSIM loss and then compare them with that without geometric constraint. From the quantitative results shown in Table 2, we can on one hand demonstrate that such 3D scene geometry-aware constraint described by a photometric loss and a SSIM loss is always helpful as it leads to a better performance on prediction accuracy for all scenes. On the other hand, this improvement from 3D scene geometry-aware constraint is applicable to different network architectures. And theoretically we can employ it to any other camera localization networks to help learning process converge towards global minimization during training. Besides, even with one Euclidean loss alone, the results prove that our proposed method performs better than PoseNet2 [4] optimized by geometric loss.

Table 2: comparison of median localization error with different network and loss terms of network on 7-Scene dataset

| Network | PoseNet [2] | | Ours | |
|---|---|---|---|---|
| Loss | $L_D$ | $L_D + L_P + L_S$ | $L_D$ | $L_D + L_P + L_S$ |
| Chess | 0.32, 8.12° | 0.11, 5.11° | 0.10, 5.38° | 0.09, 4.39° |
| Fire | 0.47, 14.4° | 0.24, 11.0° | 0.26, 13.3° | 0.25, 10.79° |
| Heads | 0.29, 12.0° | 0.16, 11.8° | 0.16, 12.6° | 0.14, 12.56° |
| Office | 0.48, 7.68° | 0.20, 8.11° | 0.22, 8.07° | 0.17, 6.46° |
| Pumpkin | 0.47, 8.42° | 0.18, 4.83° | 0.22, 6.80° | 0.19, 5.91° |
| RedKitchen | 0.59, 8.64° | 0.24, 7.19° | 0.23, 8.53° | 0.21, 6.71° |
| Stairs | 0.47, 13.8° | 0.29, 10.2° | 0.30, 11.5° | 0.26, 11.51° |

### E. Influence of depth sparsity

In indoor 7-scene dataset, dense depth maps are available directly from depth sensor. While in an outdoor environment depth information from other type of depth sensors like LIDAR or depth computation algorithms like stereo-like methods is usually sparse. Therefore, we discuss the influence of depth sparsity on our method and show that the proposed approach still works well even with a sparsity of only 20% depth information. We evaluate this property on 7-scene dataset and the results are shown in Table 3. The original depth map generated from Kinect sensor are assumed as 100% depth information. We randomly eliminate 40% of depth and 80% of depth respectively from the initial map, and then test our method using the remaining depth information without changing other network settings.

Table3: comparison of median localization error with different levels of depth sparsity

| scene | 20%-depth | 60%-depth | 100%-depth |
|---|---|---|---|
| Chess | 0.10, 5.01° | 0.10, 4.76° | 0.09, 4.39° |
| Fire | 0.25, 12.81° | 0.25, 12.38° | 0.25, 10.79° |
| Heads | 0.16, 13.31° | 0.16, 13.45° | 0.14, 12.56° |
| Office | 0.19, 7.79° | 0.17, 6.62° | 0.17, 6.46° |
| Pumpkin | 0.21, 4.74° | 0.20, 4.84° | 0.19, 5.91° |
| RedKitchen | 0.23, 10.76° | 0.22, 10.18° | 0.21, 6.71° |
| Stairs | 0.29, 12.17° | 0.28, 12.86° | 0.26, 11.51° |

Apparently, more depth information means more constraints that will evidently lead to a more precise prediction accuracy. But our method slightly outperforms event with a sparsity of 20% depth information compared to other methods illustrated in Table 1. In summary, our method can collaborate with different kinds of depth sensors or any well-defined depth computation algorithm. and provide more accurate absolute camera pose estimation.

## F. Self-supervised learning

Different from [27], we apply self-supervised learning strategy for photometric error and SSIM loss terms at training process. This means that absolute poses of image $I_{t-1}$ and $I_t$ used for building photometric consistency constraint are both predicted by network (see Figure 1(a)). To compared it, we change the pose of image $I_t$ directly from ground truth and use it to compute relative transform between two images for further warping. From the results shown in Table 4, self-supervised learning strategy outperforms both on rotation and translation accuracy. This is partly because we take more advantages of data at training process with self-supervised learning by back-propagating it twice when it is considered as $I_{t-1}$ and also when we treat it as $I_t$. Furthermore, it helps network to learn in a more natural way because camera poses are never independent to each other and for more overlapped images their corresponding poses are highly relevant by the nature of 3D geometry.

Table 4: comparison of median localization error with different learning strategy

| scene | Self-supervised learning | |
|---|---|---|
| | w/o | w |
| Chess | 0.10, 5.26° | 0.09, 4.39° |
| Fire | 0.26, 11.5° | 0.25, 10.79° |
| Heads | 0.16, 13.3° | 0.14, 12.56° |
| Office | 0.18, 7.28° | 0.17, 6.46° |
| Pumpkin | 0.25, 6.34° | 0.19, 5.91° |
| RedKitchen | 0.23, 7.53° | 0.21, 6.71° |
| Stairs | 0.29, 12.8° | 0.26, 11.51° |

## G. Outdoor evaluation on Oxford Robotcar Dataset

Our method is also tested in an outdoor dataset Oxford robotcar. Although training subset 2014-05-14-13-59-05 and test subset 2014-05-14-13-53-47 are both captured on the same day, large illumination change between two sequences and motion blur make it very challenging for camera localization.

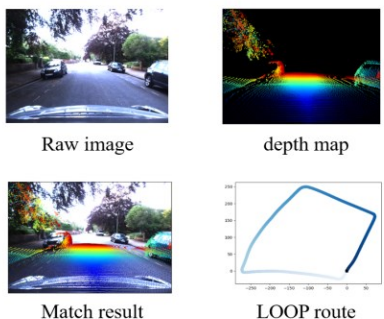

Figure 3: Oxford robotcar dataset raw color image and depth map captured by LIDAR. LOOP route subset with a total length of 1120m.

We firstly align LIDAR with a frontal camera to obtain a sparse depth map for image (see Figure 3). To avoid introducing too much depth error to our system, we choose only nearby 3D points that are less than 20m viewed from camera. The results show that our method significantly outperforms PoseNet (see Table 5) that is learned on training subset using the same network and hyper-parameters setting as [2]. Even utilizing Euclidean distant loss term alone, our method shows an accuracy increase of 15%. After introducing proposed 3D scene geometry-aware constraint, our approach provides an accuracy increase of more than 36% compared to the baseline PoseNet. To be noted that current depth map has a sparsity of less than 5% and it is also suffered from alignment noise.

To sum up, our method is not sensible to environments and it provides an apparent accuracy improvement even with highly sparse depth information. All these properties make the proposed approach suitable for many applications including indoor robots and outdoor autonomous vehicles.

Table 5: comparison of median localization error with different algorithm

| Test subset | PoseNet [2] | Ours ($L_D$) | Ours ($L_D+L_P+L_S$) |
|---|---|---|---|
| 2014-05-14-13-53-47 | 25.59, 15.96° | 22.09, 10.60° | 16.28, 7.17° |

## V. CONCLUSION

In this paper, we present a novel absolute camera localization algorithm. Rather than building a map whose size is linearly proportional to the scene size, we train a neural network to describe the scene. At the same time, we impose a novel 3D scene geometry-aware constraint as loss terms to supervise the network training. We believe that such network is more representative about 3D scene, motion and image information. The experimental results also show that our method outperforms prior works. Besides, our comparison results illustrate that positive impact is achieved when this 3D scene geometry-aware constraint is added into different networks. Therefore, we believe the effectiveness of this constraint in absolute camera localization algorithms. Last but not the least, our method is suitable for many applications like indoor mobile robots or outdoor autonomous driving vehicles. On these platforms, training data is directly available from different sensors and no additional manual annotation is required. In future work, we aim to pursue further fusion between CNN-based methods and traditional metric-based methods for camera localization. And an integration of different sensor modalities may also improve camera localization.